\documentclass[letterpaper, 10 pt, conference]{ieeeconf}  

\IEEEoverridecommandlockouts                              

\usepackage{graphics} 
\usepackage{epsfig} 
\usepackage{times} 
\usepackage{amsmath} 
\usepackage{amssymb}  
\usepackage[T1]{fontenc}  
\usepackage{bm}
\usepackage{url}
\usepackage{float}
\usepackage{multirow}
\usepackage[noadjust]{cite}
\usepackage{dblfloatfix} 

\title{\LARGE \bf Applying Extended Object Tracking for Self-Localization of Roadside Radar Sensors}

\author{Longfei Han$^{1,2,\dagger}$, 
Qiuyu Xu$^{1,\dagger}$, 
Klaus Kefferp\"utz$^{1,3}$, 
Gordon Elger$^{1,3}$,
and J\"urgen Beyerer$^{2,4}$
\thanks{$^{1}$Application Center \guillemotright Connected Mobility and Infrastructure\guillemotleft , Fraunhofer IVI, Ingolstadt, Germany}
\thanks{$^{2}$Vision and Fusion Laboratory, Karlsruhe Institute of Technology (KIT), Karlsruhe, Germany}
\thanks{$^{3}$Technische Hochschule Ingolstadt, Ingolstadt, Germany}
\thanks{$^{4}$Fraunhofer IOSB, Karlsruhe, Germany}
\thanks{$^\dagger$These authors contribute equally in this work.}
\thanks{Corresponding author: {\tt\small longfei.han@ivi.fraunhofer.de}}
}

\begin{document}

\maketitle
\thispagestyle{empty}
\pagestyle{empty}

\begin{abstract}

Intelligent Transportation Systems (ITS) can benefit from roadside 4D mmWave radar sensors for large-scale traffic monitoring due to their weatherproof functionality, long sensing range and low manufacturing cost.
However, the localization method using external measurement devices has limitations in urban environments.
Furthermore, if the sensor mount exhibits changes due to environmental influences, they cannot be corrected when the measurement is performed only during the installation.
In this paper, we propose self-localization of roadside radar data using Extended Object Tracking (EOT).
The method analyses both the tracked trajectories of the vehicles observed by the sensor and the aerial laser scan of city streets, assigns labels of driving behaviors such as "straight ahead", "left turn", "right turn" to trajectory sections and road segments, and performs Semantic Iterative Closest Points (SICP) algorithm to register the point cloud.
The method exploits the result from a down stream task -- object tracking -- for localization.
We demonstrate high accuracy in the sub-meter range along with very low orientation error.
The method also shows good data efficiency.
The evaluation is done in both simulation and real-world tests.
\end{abstract}

\section{Introduction}
Using 4D automotive radar sensors in Intelligent Transportation Systems (ITS) could benefit large scale traffic analyses from to its relatively simple mechanical design and low manufacturing cost. 
They provide 3D point clouds with radial velocity attributes over a long range in adverse weather conditions. 
However, the localization of such sensors in ITS sensors is a challenging task \cite{localizationiv}. 
On the one hand, localization approaches using external measurement devices (e.g. Global Navigation Satellite System GNSS) have limitations in urban environments due to the multi-path problem and the non-light-of-sight effect. 
On the other hand, localization methods using sensor data of targets are limited by the current low resolution and thus the difficulty of  finding known features in the environment (e.g. lantern poles or buildings).

Given the application of ITS, traffic is both the main interest and the largest feature at the same time. 
We propose to localize the radar sensors by comparing their traffic measurements  with the road in representations such as the map and the aerial laser scan.
More specifically, we perform object tracking using the radar sensor measurements, forming trajectories of the objects, and matching these trajectories to the geometry of roads obtained from aerial laser scan.
A novelty in the pipeline is the use of an extended object model \cite{eot} to preserve the shape description of the vehicle, which is further enriched to provide a better shape description of the complete trajectories.
When collecting abundant trajectories, they represent the road geometry and form the counterpart for  localization using algorithms such as  iterative closest points (ICP) \cite{icp} and solve the localization problem through point cloud registration. 
Among the Extended Object Tracking (EOT) methods  (for an overview see \cite{eot}), a star-convex model \cite{StarConvex} is chosen for its potential to represent any free-form object.
We further implement the EOT based on Gaussian process approaches \cite{GP, GPimproved}.
The resulting trajectories from the tracker are represented with point clouds.
They are labelled with different geometric features as well as driving behaviors like "straight ahead", "left turn", "right turn" to obtain a finer lane-level details, which benefits the point registration work.
To cope with the labeled point clouds, a semantic ICP (SICP) \cite{sicp} approach is implemented, so that this information is taken into consideration.

Our contributions are summarized as follows:
\begin{enumerate}
    \item We propose a novel self-localization method for ITS roadside radar sensors. 
This approach treats the localization as a point cloud registration problem.
The source and target point clouds are labelled and processed with the SICP algorithm.
EOT algorithms are used to generate the source point cloud.
Publicly available aerial laser scan (ALS) point clouds are used as the target point cloud.
    \item  We demonstrate this method in both simulative and real-world scenarios and show improvements compared to other methods.
\end{enumerate}

The remainder of the paper begins with a review of related works in radar sensor localization and EOT in Section \ref{Chapter2: RelatedWorks}. 
Section \ref{Chapter3: ProblemFormulation} formulates the problem as an object tracking and point cloud registration task. 
Section \ref{Chapter4: Implementation} introduces the methods.
We show how the building blocks such as  EOT, SICP are assembled to form the self-localization pipeline.
Following this, experiments and validations are elaborated in Section \ref{Chapter5: Evaluation}.
Section \ref{Chapter6: Conclusion} concludes this work and shows future works.

   \begin{figure*}[t]
      \centering
      \includegraphics[scale=0.5]{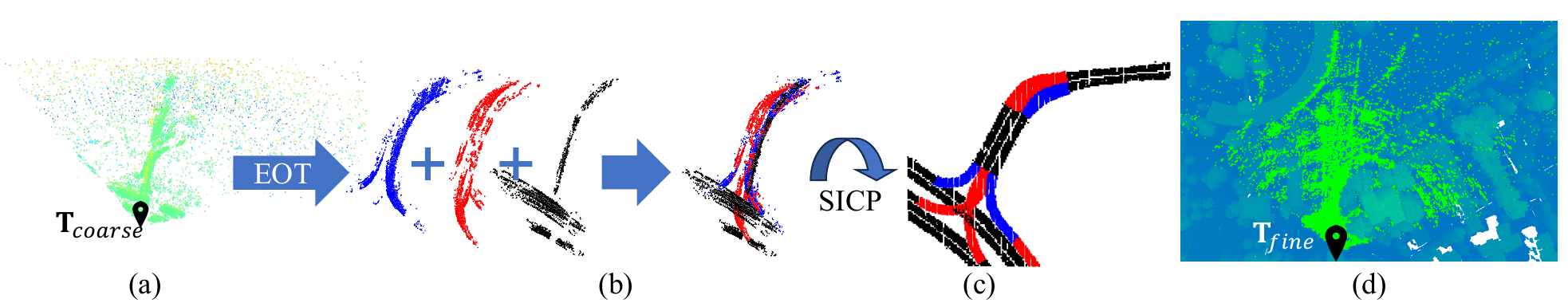}
      \caption{Conceptual overview of the proposed localization method with trajectories collected using EOT. (a) Radar raw data and a coarse transformation matrix $\mathbf{T}_{coarse}$.
      (b) Accumulation of trajectories from different vehicle using EOT tracker.
      Blue, red and black are the "right turn", "left turn", "straight ahead" trajectories.
      (c) Preprocessing of aerial laser scan data to get labelled lanelets. 
      Colors also represent driving behaviors.
      HD maps are used to crop out the road from the aerial laser scan.
      SICP algorithm is used to register the source point cloud to the target point cloud.
(d) The resulting $\mathbf{T}_{fine}$ is used to calculate the sensor pose $\mathbf{T}_{\texttt{UTM}}$.
Light green points are sensor data.
Dark green and blue points are the aerial laser points.
      }
      \label{Figure:01_Concept}
   \end{figure*}
\section{Related Works}
\label{Chapter2: RelatedWorks}
Sensor localization is an active research area. 
Many works address this topic for radar sensors.
Most of them focus on the radar sensor on a moving base \cite {RSL-Net, overhead, RadarLoc, radarosm}.
Few focuses on roadside static sensors \cite{localizationiv}. 
An intuitive approach is to localize the radar sensor with geodata. 
OpenStreetMap (OSM) serves as a community-supported data source and enables its application in large-scale localization tasks. 
Beside \cite{OSMLaser, OSMlidar}, where OSM maps are used for lidar localization, Hong et al. extract surface features from radar polar images and match these with surrounding map semantics \cite{radarosm}.

Aerials imagery has also been used for radar localization task.
One example is the approach introduced in \cite{RSL-Net}, for the localization of a Frequency-Modulated Continuous-Wave (FMCW) radar mounted on a moving vehicle. 
An overhead imagery was used to refine the coarse initial estimate from the odometry or Global Navigation Satellite System (GNSS) devices.
This approach was extended to a self-supervised fashion to simplify the training process while maintain the accuracy \cite{overhead}.
Another learning based approach has been seen for radar relocation task \cite{RadarLoc}.
The authors trained a self-attention mechanism to directly regress the 6D-pose in a global frame.
Another approach to localize a radar sensor is to find its pose in a global lidar map, given the higher density of the points \cite{radaronlidar, RoLm}. 
The methods above mainly address moving sensors. 
Theoretically, an adaptation to ITS radar sensors is possible, the results though are in the range of meters and degrees not adequate for ITS applications in case of an object from a large distance.
A novel approach \cite{localizationiv} address the localization of ITS roadside radar sensors using aerial laser scan and accumulated dynamic point cloud from a 4D mmWave radar. 
This work manages to localize roadside radar sensor in a self-localization manner but does not provide road information in the lane level. 

Using vehicle trajectories to localize the radar sensor means also to extract road information from sensor data.
Koch et al. used Ground Target Moving Indicator (GTMI) radar to track the vehicles on the ground with Multi-Hypotheses Tracking (MHT) and use the trajectory for road extraction \cite{MapExtraction}. 
As an example of sensors with high resolution, lidar point clouds are used to generate lane information using background subtraction \cite{lidarRoad}. 
No tracking is used yet in this work. 
Extended object tracking is another active research area and has seen a lot of developments \cite{eot}.
It could be mainly divided into ellipses shape \cite{RandomMatrix} and star-convex shapes \cite{StarConvex}.
The latter is more suitable for traffic participants descriptions given its ability to describe free form. 
Some algorithms for tracking star-convex objects with Gaussian Process (GP) show good results in both 2D and 3D scenarios \cite{GP,GPimproved, GP3D}.
It enables shape descriptions of vehicles and other traffic participants and use the shape information to cumulatively describe the road with lane-level information.
Yet an approach for road extraction with EOT has not been seen.

\section{Problem Formulation}
\label{Chapter3: ProblemFormulation}
The focus of this work is to determine the pose of the radar sensor in a Universal Transverse Mercator (UTM, e.g. UTM32N for Bavaria, Germany) coordinate system. 
The pose is represented as a transformation matrix $\mathbf{T}_{\texttt{UTM}}$, with
\begin{equation}
\mathbf{T}_{\texttt{UTM}} = 
\begin{pmatrix}

    \mathbf{R}_{\texttt{UTM}}^{SO(3)} & \bm{t}_{\texttt{UTM}} \\
    \begin{matrix} 0 & 0 & 0 \end{matrix} & 1
\end{pmatrix}.
\end{equation}
$\mathbf{R}_{\texttt{UTM}}^{SO(3)} \in SO(3)$ is the rotation matrix of the reference frame of the sensor with respect to the East-North-Up global frame and could also be reduced to $\begin{pmatrix}
  \mathbf{R}_{\texttt{UTM}}^{SO(2)} & \bm{0}\\ 
  \bm{0} & 1
\end{pmatrix}$, when the pitch and roll angles are small and could be neglected.
This includes the situation where the sensor are mounted parallelly to the ground to achieve a large range of detection.  
$ \bm{t}_{\texttt{UTM}} = (x_{\texttt{UTM}}, y_{\texttt{UTM}}, z_{\texttt{UTM}})^T$ is the translational vector. 

In order to achieve the transformation matrix, we follow a three-step procedure (Fig \ref{Figure:01_Concept}).
The first step is to perform the coarse localization presented in \cite{localizationiv}.
The coarse localization result transfers the sensor point cloud into a near flat plane for the 2D tracking methods.
The result is $\mathbf{T}_{coarse} \in SE(3)$.
In the second step we use extended object tracking with the radar sensor data to form trajectories of the vehicles.
The third step is the fine localization by registration of point cloud representing the vehicle trajectories to the point cloud of the aerial laser scan.
Even though Özkan et al. showed the 3D version of the Gaussian Process \cite{GP3D} based extended object tracking, directly implementing this approach is out of scope of this paper and could be explored in following works.
The third step results in $\mathbf{T}_{fine} \in SE(2)$. 
To achieve the final result, we extend $\mathbf{T}_{fine} \text{ to } \tilde{\mathbf{T}}_{fine}\in SE(3)$, and have $\mathbf{T}_{\texttt{UTM}} = \tilde{\mathbf{T}}_{fine}\mathbf{T}_{coarse}$.
\subsection{Self-localization with accumulated dynamic point clouds}
\label{Chapter:IV2024}
In \cite{localizationiv}, a method for self-localization is introduced using chronically accumulated dynamic point clouds $P^d = \{(x,y,z,\dot{r}) \mid \dot{r} \geq \lambda\}$, where $\lambda$ is the range rate threshold to extract the "moving" points, and aerial laser scan point cloud  $P_{\texttt{UTM}}^r = \{(x_{\texttt{UTM}},y_{\texttt{UTM}},z_{\texttt{UTM}}) \mid (x_{\texttt{UTM}},y_{\texttt{UTM}},z_{\texttt{UTM}}) \in \text{road}\}$. 
Note $P^d$ are in the local reference frame of the sensor and  $P_{\texttt{UTM}}^r$ is in the UTM coordinate system.
Comparing to \cite{localizationiv}, we drop the Kalman Filter part and use the method only as an initialization of the transformation matrix. 
Particularly, for $\bm{p}^d \in P^d$ and  $\bm{p}_{\texttt{UTM}}^r \in P_{\texttt{UTM}}^r$, we use ICP with different range to calculate the coarse transformation matrix $\mathbf{T}_{coarse}$ (see Fig. \ref{Figure:01_Concept}(a)), so that
\begin{equation}
\mathbf{T}_{coarse} = \underset{\mathbf{T}}{\text{arg min}} \, \sum^{n} \|\mathbf{T}_{coarse} \bm{p}^d - \bm{p}_{\texttt{UTM}}^r\|^2
\end{equation}
For more details, please refer to \cite{localizationiv}.

\subsection{Extended Object Tracking using Gaussian Processes}
Following \cite{GP}, we are interested in the vehicles' states $\bm{x}_{\texttt{UTM}}$ consisting of the kinematic features $\bm{x}_{\texttt{UTM}}^k$ and the contour feature $\bm{x}^c$:
\begin{equation}
\bm{x}_{\texttt{UTM}} = [\bm{x}_{\texttt{UTM}}^{k^T}, \bm{x}^{c^T}]^T.
\end{equation}
For $\bm{x}_{\texttt{UTM}}^k$, the 2D position ($x_{\texttt{UTM}}, y_{\texttt{UTM}}$) on the east-north plane along with the speed $v$, the longitudinal acceleration $a$, and the yaw angle and the yaw rate ($\phi, \dot{\phi}$) are required, corresponding to the chosen motion model, constant turn rate and acceleration (CTRA). 
\begin{equation}\label{eq:StateVector}
\bm{x}_{\texttt{UTM}}^k = [x_{\texttt{UTM}}, y_{\texttt{UTM}}, v, a, \phi, \dot{\phi}]^T.\end{equation}
The contour is described using the star-convex model \cite{StarConvex} and further parameterized in a polar coordinate system using a radial function $d = f(\theta)$, which forms the contour feature $\bm{x}^c$.
Given the  $n_\theta$ angles $\bm{\theta} = [\theta_1, \theta_2, \ldots, \theta_{n_\theta}]$ and the distances between a corresponding point on the contour and the center of the object $D=\{f(\theta_1), f(\theta_2), ..., f(\theta_{n_\theta})\}$, we have 
\begin{equation}
\bm{x}^c = [f(\theta_1), \ldots,  f(\theta_{n_\theta})]^T.
\end{equation}

Combined with the object center $(x_{\texttt{UTM}}, y_{\texttt{UTM}})$, the contour is represented as:
\begin{equation}
\bm{c}_{\texttt{UTM}, i} = \begin{bmatrix}           x_{\texttt{UTM}} \\ y_{\texttt{UTM}}\end{bmatrix} + \begin{bmatrix}           \cos(\theta_i) \\ \sin(\theta_i)\end{bmatrix} f(\theta_i) + \bm{e}_i, \forall i, 1 \leq i \leq n_{\theta}.
\end{equation}

The GP is then used to approximate $f(\cdot)$, as 
\begin{equation}\label{eq:GP}
f(u) \sim \mathcal{GP}(\mu(u), k(u, u')),\end{equation}
where
\begin{subequations}
\begin{align}
 \mu(u) &= \mathbb{E}[f(u)], \\ 
k(u,u') &= \mathbb{E}[(f(u) - \mu(u))(f(u')-\mu(u'))^T],
\end{align}
\end{subequations}
$u$ and $u'$ are the input of function $f(\cdot)$.
Note the GP indicates the normal distribution of the function values evaluated at finite input $u_1, \ldots, u_{n}$, so Eq. (\ref{eq:GP}) becomes
\begin{equation}
\begin{bmatrix} f(u_1)\\\vdots\\f(u_n)  \end{bmatrix} = \mathcal{N}(\bm{\mu},\mathbf{K}),
\end{equation} with 
\begin{equation}
\bm{\mu} = \begin{bmatrix} \mu(u_1)\\ \vdots \\\mu(u_n)  \end{bmatrix},
\mathbf{K} = \begin{bmatrix} k(u_1,u_1) & \cdots & k(u_1, u_N)\\\vdots &\ddots& \vdots\\k(u_N,u_1) & \cdots & k(u_N,u_N)  \end{bmatrix}.
\end{equation}
\begin{equation}
    k(u,u') = \sigma^2_f \cdot e^{-\frac{2 \sin(\frac{\vert u-u' \vert}{2} )^2 }{l ^2}} + \sigma^2_r,
\end{equation} 
as $u \in [0,2\pi)$ in this setup. 
$\sigma_f$ is the prior standard deviation, $\sigma_r$  is the radius standard deviation, and $l$ is the length scale.
The GP regression in the literature is performed in a recursive manner to be solved with a Kalman Filter (e.g. \cite{GP}). 
The motion model is defined for $\bm{x}_{\texttt{UTM}}^k \text{ and } \bm{x}^c$ independently.
The motion model for  $\bm{x}_{\texttt{UTM}}^k$ is CTRA  \cite{CTRA}. 
For $\bm{x}^c$, we have
\begin{equation}\bm{x}^c_{t+1} = e^{-\tau\Delta t} \cdot \bm{x}^c_{t}, \mathbf{Q^c_t} = (1- e^{-2\tau\Delta t} ) \cdot \mathbf{K}\end{equation}
where $\tau$ is a non-negative forgetting factor,  $\mathbf{Q}^c_{t}$  is the covariance matrix.
$\mathbf{K}$ (below $\mathbf{K}(\bm{\theta}, \bm{\theta})$)  denotes the covariance matrix of the basis point angles and it does not change in the process.
Following $\bm{x}^c = \bm{f}_{n_\theta \times 1}$, when a new measurement is received from the angle $\theta^\ast$, we have
\begin{equation}
    \begin{bmatrix}\bm{f} \\ f^\ast  \end{bmatrix} = \mathcal{N}
    \begin{pmatrix}
        \bm{0}, 
        \begin{bmatrix}
            \mathbf{K}(\bm{\theta}, \bm{\theta}) & \bm{k}(\bm{\theta}, \theta^\ast) \\
            \bm{k}(\theta^\ast , \bm{\theta}) & k(\theta^\ast ,\theta^\ast)
        \end{bmatrix} 
    \end{pmatrix}
,\end{equation}
and 
\begin{equation}\label{eq:measurement}
    \begin{split}
    f^\ast \vert \bm{f} \sim \mathcal{N}( &\bm{k}(\theta^\ast , \bm{\theta})\mathbf{K}(\bm{\theta}, \bm{\theta})^{-1}\bm{f},  \\
    & k(\theta^\ast ,\theta^\ast) - \bm{k}(\theta^\ast , \bm{\theta})\mathbf{K}(\bm{\theta},\bm{\theta})^{-1}\bm{k}(\bm{\theta}, \theta^\ast ))
    \end{split}
.\end{equation}
The measurement model is derived from Eq. (\ref{eq:measurement}).
In order to have the measurements better associated to $\theta^\ast$, the sensor position should be taken into consideration \cite{GPimproved}.
It involves the generation of many visible candidates of reflection spots under the consideration of the relative position between the sensor and the vehicle, and the association of real measurements to the candidates with a Hungarian Algorithm. 
The detailed elaboration is out of scope for this work, we refer readers to \cite{GP, GP3D, GPimproved}.
The output of this extended tracking are the states of all tracked vehicles $X_{\texttt{UTM}}$  and the measurements $Z_{\texttt{UTM}} \subseteq P^d_{\texttt{UTM}}$, which are used at different time instance for the update step in the filter.
This step is depict in Fig. \ref{Figure:01_Concept}(b).
\subsection{Semantic Iterative Closest Point}
\label{Chapter:sicp}
We use the semantic iterative closest point (SICP) \cite{sicp} for the point cloud registration task. 
The algorithm aligns the selected measurements point cloud $Z_{\texttt{UTM}}$ (source point cloud) to the road $P_{\texttt{UTM}}^r$ (target point cloud), represented in Fig. \ref{Figure:01_Concept}(b) and Fig. \ref{Figure:01_Concept}(c) respectively.
The semantic is represented by labels of 3 categories: $S \triangleq \{s_k \mid s_k \in \{\text{left turn}, \text{right turn}, \text{straight ahead}\}\}$, using angular curvature which results from yaw rate and speed of the vehicle states $X_{\texttt{UTM}}$ and the curvature of the road segments and their corresponding thresholds.  
We adopt SICP and consider these semantic labels of the point clouds in the joint probability $p(R, S, I \mid P_{\texttt{UTM}}^r,  Z_{\texttt{UTM}})$, where 
\begin{equation}
\begin{split}
  R=\{\bm{r}_i = &\bm{p}_{\texttt{UTM},i}^r - \mathbf{T}\bm{z}_{\texttt{UTM},i} \mid \\ &\bm{p}_{\texttt{UTM},i}^r \in P_{\texttt{UTM}}^r, \bm{z}_{\texttt{UTM},i} \in Z_{\texttt{UTM}}, \mathbf{T} \in SE(2)\}    
\end{split}   
\end{equation}
is the residual between target point cloud and transformed source point cloud, and $I$ is the association.
The result $\mathbf{T}_{fine}$ (Fig. \ref{Figure:01_Concept}(d))can be calculated by solving the \textit{maximum likelihood estimation} (MLE) problem:
\begin{equation}
\mathbf{T}_{fine} = \underset{\mathbf{T} \in SE(2)}{\text{arg max}} \, f(\mathbf{T};R, S, I \mid P_{\texttt{UTM}}^r,  Z_{\texttt{UTM}}),
\end{equation}
where
\begin{equation}    
f(\mathbf{T};R, S, I \mid P_{\texttt{UTM}}^r,  Z_{\texttt{UTM}}) \triangleq \log P(R, S, I \mid P_{\texttt{UTM}}^r,  Z_{\texttt{UTM}}).
\end{equation}  

\section{Implementation}
\label{Chapter4: Implementation}

The main procedure of our localization method is to get the target and source point clouds for the point cloud registration.
In this section, we describe in detail the generation of these point clouds for two processes, namely Sections \ref{Chapter:IV2024} and \ref{Chapter:sicp}.

\subsection{Generation of Target Point Cloud}
The target point cloud results from the aerial laser scan \cite{LaserScan}.
In other areas, local authorities are also providing similar resources.
The laser scan of the roads are the target point cloud. 
In order to get these, a semantic segmentation of the point cloud is needed.
Here we use manual labeling, with the help of HD Map format Lanelet2 \cite{lanelet2o}. 
From aerial image, HD maps of the region of interest (ROI) are drawn.
We then separate the aerial laser scan using the left and right bounds of the lanelets.
These point clouds further undergo through down sampling process like voxel grid calculation.
The resulting aerial laser scan can be cropped to the adequate range to help keep the fitness of the ICP process in a reasonable range. 
This $P_{\texttt{UTM}}^r$ is necessary for Section \ref{Chapter:IV2024}.
For SICP, the elements in $P_{\texttt{UTM}}^r$ need to have labels $S$. 
We extend the $\bm{p}_{\texttt{UTM}}^r = (x_{\texttt{UTM}},y_{\texttt{UTM}},z_{\texttt{UTM}}) $ to $\bm{p}_{\texttt{UTM},s}^r = (x_{\texttt{UTM}},y_{\texttt{UTM}},z_{\texttt{UTM}}, s)$.
$s$ is defined by the curvature of the lanelet.
Given the set of lanelet $L$, we select the left bound of $l_i \in L$, and find its start point, middle point and the end point $\bm{p}_{st}^{l_i},\bm{p}_{md}^{l_i},\bm{p}_{ed}^{l_i} $ , then calculate the Menger curvature, 
\begin{equation}
\begin{split}
    c &= f(\bm{p}_{st}^{l_i},\bm{p}_{md}^{l_i},\bm{p}_{ed}^{l_i}) \\ &= \frac{4 \mathrm{Area}(\bm{p}_{st}^{l_i},\bm{p}_{md}^{l_i},\bm{p}_{ed}^{l_i})}{\Vert \bm{p}_{st}^{l_i}-\bm{p}_{md}^{l_i}\Vert \cdot \Vert  \bm{p}_{md}^{l_i}-\bm{p}_{ed}^{l_i}\Vert \cdot \Vert  \bm{p}_{ed}^{l_i}-\bm{p}_{st}^{l_i} \Vert},
\end{split}
\end{equation}
which is to be compared with a threshold $\gamma$, to get
\begin{equation}
    s_i = \begin{cases}
        \text{left turn}, & \text{for } c > \gamma \text{ and } \bm{p}_{st}^{l_i},\bm{p}_{md}^{l_i},\bm{p}_{ed}^{l_i} \\ 
        & \text{are arranged counter-clockwise}\\
        \text{right turn}, & \text{for } c > \gamma \text{ and } \bm{p}_{st}^{l_i},\bm{p}_{md}^{l_i},\bm{p}_{ed}^{l_i} \\
        & \text{are arranged clockwise}\\
        \text{straight}, & \text{for } c \leq \gamma.\\
        \end{cases}
\end{equation}
Here $\mathrm{Area}(\bm{p}_{st}^{l_i},\bm{p}_{md}^{l_i},\bm{p}_{ed}^{l_i})$ is the area of the triangle formed by the points.
$\gamma$ is 0.01$m^{-1}$.
For  $\bm{p}_{\texttt{UTM},s}^r$ that falls in $l_i$, $s=s_i$.

\subsection{Generation of Source Point Cloud}
The source point cloud originates from the radar sensor data. 
For the coarse point cloud registration using all points with a radial doppler reading larger than a threshold, we accumulate the point cloud to a predefined number of frames and filter out the points with low doppler readings.
The source point cloud further undergoes a voxel-based down sampling process and a DBSCAN \cite{dbscan} filter so that noisy outliers are rejected for further processing.
A coarse trimming to fit the observing area could be added optionally, in order to have reasonable high fitness in the ICP process.
We refer to \cite{localizationiv} for more detailed description.

The coarse ICP process is then performed and the radar point cloud is transformed using the result.
By doing this, the possible pitch and roll angles of the sensor with respect to the street surface could be largely ruled out, so that further processing could be conducted using an algorithm designed for 2D measurements. 
As mentioned in Section \ref{Chapter3: ProblemFormulation}, Gaussian Process Extended Object Tracking (GP-EOT) is used to find trajectories of the vehicles along with their sensor measurements. 
The GP-EOT algorithm starts with a defined birth place of objects, commonly at the spots where the edges of the field of view and the streets intersect. 
Each vehicle that drives passing this spot will be tracked individually. 
The spot is also used in the prior information for the calculation of the position in the kinematic states (\ref{eq:StateVector}).
The tracker then updates the kinematic states using an unscented Kalman Filter \cite{UKF}. 
The shape information in this work is represented using 20 basis point angels ($n_{\theta}=20$) in the shape vector $\bm{x}^c$. 
Different trajectories are output of the tracker instances.
The measurements which are used to generated these trajectories are accumulated.
So is the shape information.
We use yaw rate and speed to calculate the angular curvature $\kappa$ of the states in the trajectory to assign a label to the point cloud.
For these points in $Z_{\texttt{UTM}}$, their labels $s_i$ is then 
\begin{equation} \label{eq:SourceLabel}
    s_i = \begin{cases}
        \text{left turn}, & \text{for } \kappa > \eta \\ 
        \text{right turn}, & \text{for } \kappa < -\eta \\
        \text{straight}, & \text{for } -\eta \ \leq \kappa \leq \eta\\
        \end{cases}, \kappa = \dot{\phi}/v
\end{equation}
$\eta$ is selected to be 0.01$rad/m$.
The labeled point cloud of radar measurements are also down sampled to the same scale of voxels to be used as source point clouds in the SICP process. 

The SICP process chooses the maximum corresponding distance between a point in the source cloud and a point in the target cloud to be 50m and the convergence criterion to be 100 iterations or changes in RMSE smaller than 1e-4.

\begin{figure}[t]
      \centering
      \includegraphics[scale=0.3]{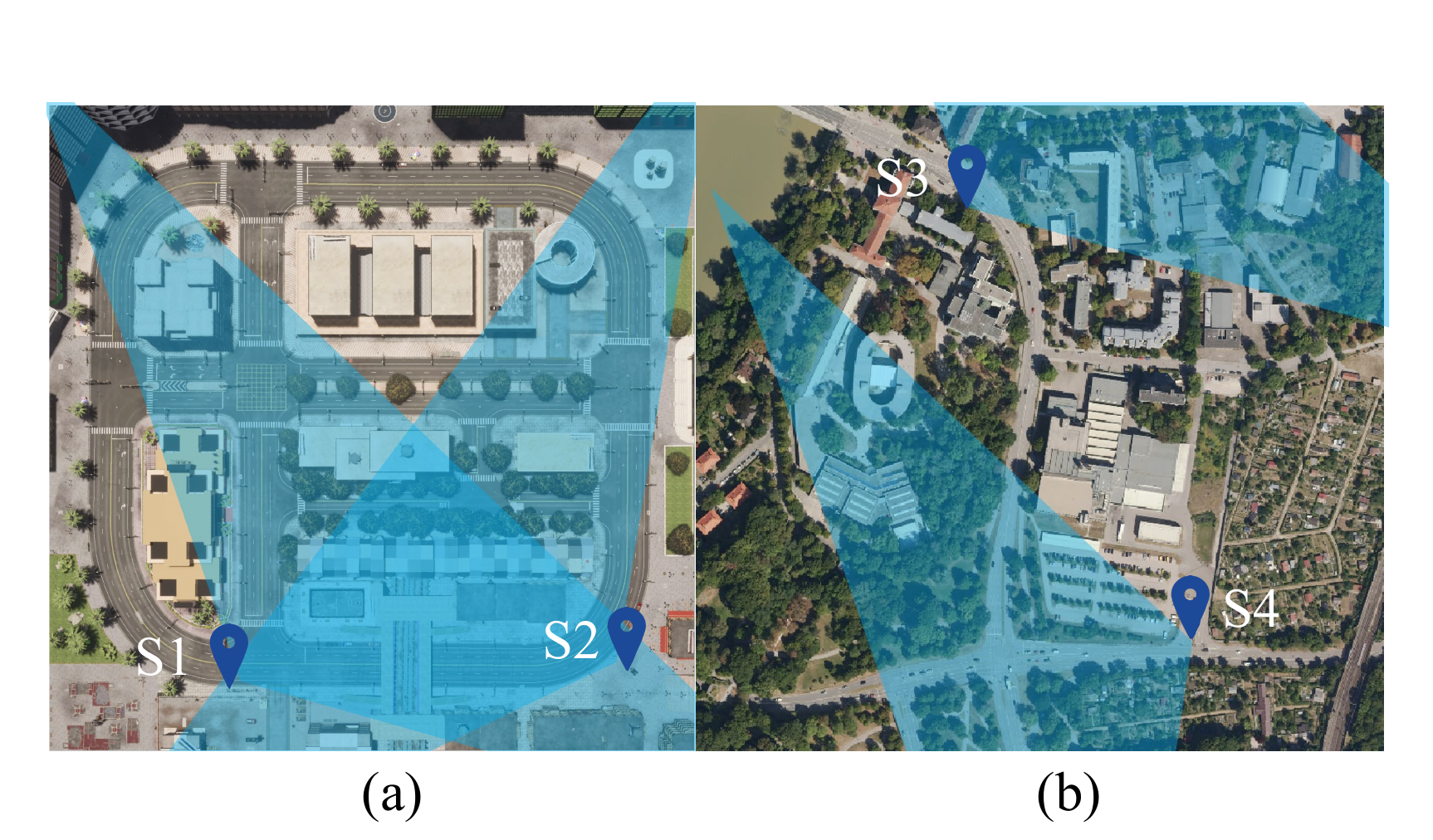}
      \caption{Test scenarios.
      (a) Sensor S1 and S2 are placed in CARLA Simulation map Town 10. 
      (b) S3 and S4 are placed in Ingolstadt Germany.
      The location symbols indicate the position of the sensor.
      The triangles depict the viewing direction and an approximate range of view.
      The image in (b) is retrieved from \cite{AerialImage}}.
      \label{Figure:Evaluation}
   \end{figure}
\section{Evaluation}
\label{Chapter5: Evaluation}

The evaluation of the method is divided into three parts.
Since the coarse localization is introduced in \cite{localizationiv}, we start with the quality of our GP-EOT implementation.
After that, the localization is evaluated using both simulative data in the CARLA simulator \cite{carla} and real world sensor data.
The third part show cases the comparison of the result from this method to others.
We evaluate the four deployment positions as shown in Fig. \ref{Figure:Evaluation}.
\begin{figure}
\centering
\includegraphics[width=0.75\linewidth]{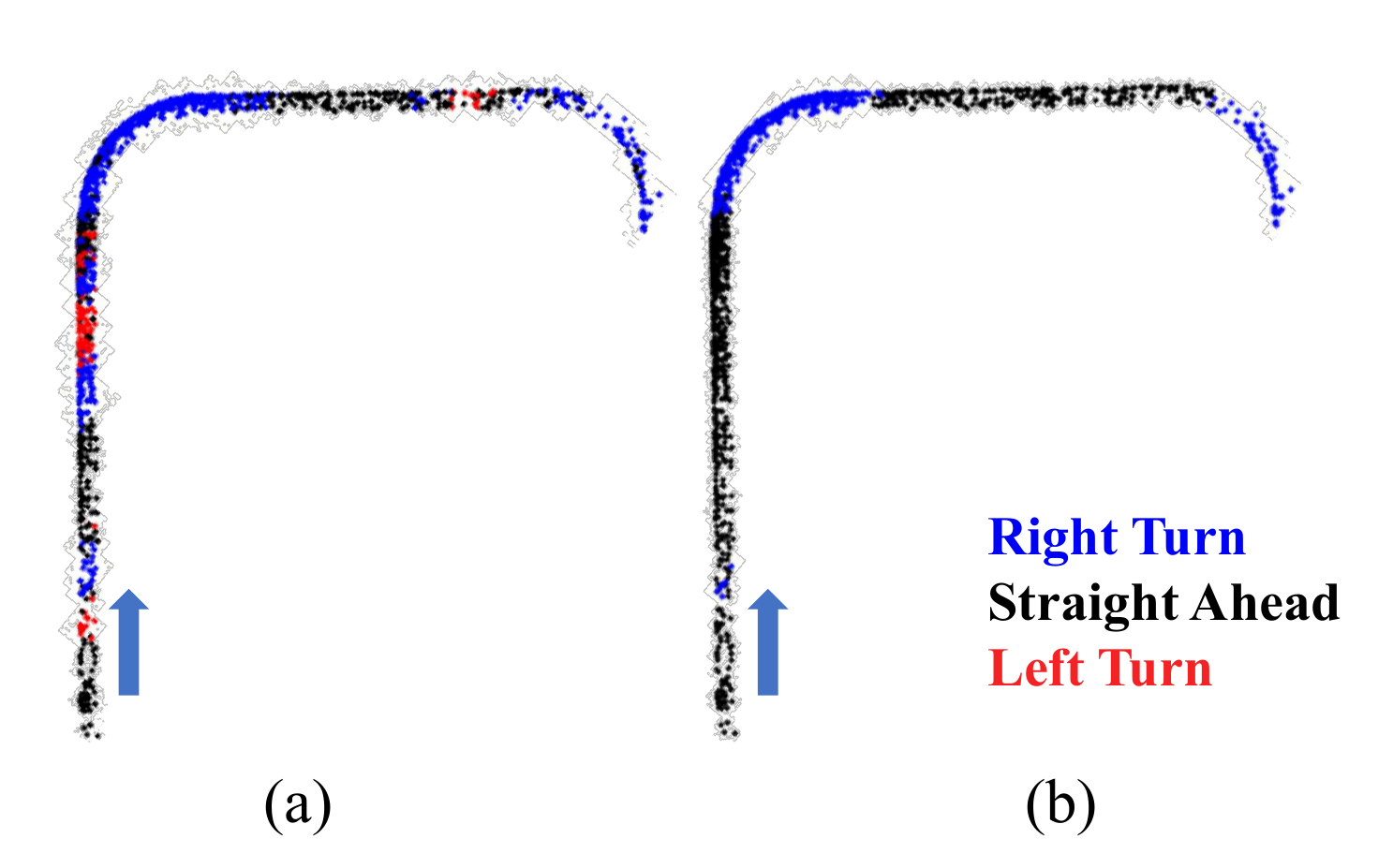}    
    \caption{Example of result from GP-EOT and labels for indicating the driving behavior of the vehicle. 
    (a) shows the result with measurement model from \cite{GP}, where the sensor position is not considered in the measurement model.
    (b) shows the result with measurement model from \cite{GPimproved}, with sensor position considered in the measurement model. 
    The color indicates the driving behavior: right turn, left turn and straight ahead. 
    The tracking starts at the position of the arrow.}
    \label{Figure:RadarLabel}
\end{figure}
\begin{table}
    \caption{Average absolute error of the localization}
    \label{tab:Result}
    \centering
    
    \begin{tabular}{cccc}
    \hline
         Sensor&  $\text{Error}_x$ (m)&  $\text{Error}_y$ (m)& $\text{Error}_{yaw}$ (°)\\
    \hline
         S1&  0.60&  0.28& 0.20\\
         S2&  \textbf{0.16}&  0.93& 0.22\\
         S3&  0.32&  0.21& 0.35\\
         S4&  0.52&  \textbf{0.19}& 0.16\\
    \end{tabular}

\end{table}
\subsection{Tracking Quality}
EOT tracking serves as the fundamental step in processing the radar sensor data to get the source point cloud for ICP. 
The quality influences the data gathering step by selecting which points should be included into the trajectory.
It further affects the labeling step with the estimated yaw rate value (see (\ref{eq:SourceLabel})).
We evaluated 60 tracked trajectories in the CARLA  simulator from the Sensor S1 in Fig. \ref{Figure:Evaluation}.
The average RMS of distance between estimated center point of vehicle and the ground truth, $d_{x,y}$, the average difference between estimated yaw angle and the ground truth, $ d_\phi$,  the Intersection over Union (IoU) of estimated and ground truth shape are 0.49m, 3.8° and 0.59, respectively. 
We also compare the result between two GP-EOT with measurement model from \cite{GP} and \cite{GPimproved}, as shown with an example trajectory in Fig. \ref{Figure:RadarLabel}.
The tracking algorithm returns clear trajectory with good labeling result, especially when considering the relative position between the sensor and the vehicle. 
The vehicle is observed from the lower part of the diagram, drives upward and turns twice to the right. 
The clear color separation shows that our implemented tracker is able to distinguish different behaviors of the vehicle.
This information can then be matched to the predefined behavior of the lane, as shown in Fig. \ref{Figure:01_Concept}(c), where different colors indicates the lane should be used for turns or straight drives.

\begin{figure*}[t]
    \centering
    \includegraphics[width=1\linewidth]{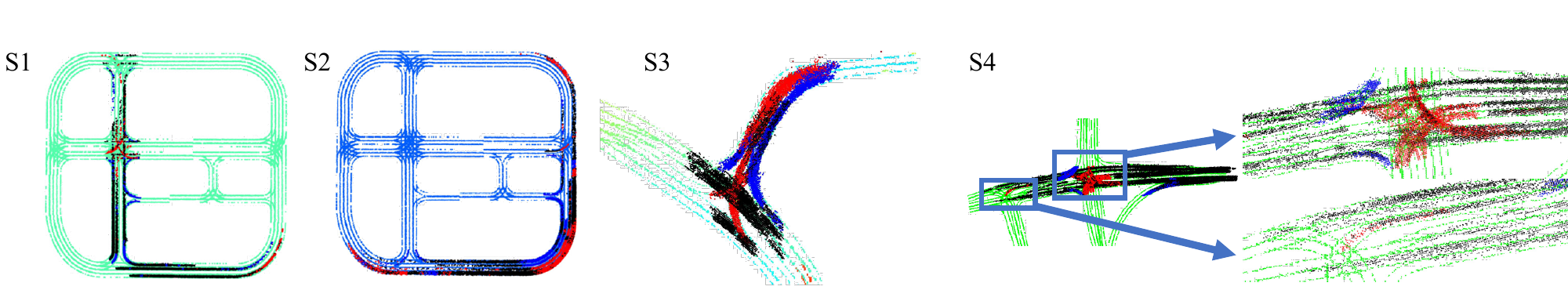}
    \caption{Localization result of the four sensors.
    Both target and source points for SICP are shown in the figure.
    For radar point clouds, the blue points indicate left turning behavior, the red points indicate right turning behavior and the black points indicate driving straight ahead.
    The bottom right part extend the two }
    \label{fig:Result}
\end{figure*}

   
\subsection{Localization of Radar Sensors}
The data for evaluating the radar sensor localization are gathered with the CARLA radar model in the simulation and a Continental ARS548 in a mobile ITS sensor setup \cite{flexsense}.
Sensor 1 and Sensor 2 are placed in CARLA Town 10.
Sensor 3 and Sensor 4 are placed in Ingolstadt, Germany. 
The ground truth is measured using GNSS device with Real-Time-Kinematics (RTK) correction data.
The sensors are all set to have 120° horizontal field of view (FOV) and 30° vertical FOV.
The points gathered are approximately 10000 points per second.
The aerial laser scan and the aerial image is retrieved from open database provided by the Agency for Digitisation, High-Speed Internet and Surveying of Bavaria \cite{LaserScan, AerialImage}.
To localize Sensor 1, 20085 points are used, 15250 are labeled with straight ahead, 3439 points are labeled as right turn, 1396 points are from left turn behavior.
To localize Sensor 2, 44511 points are used, 13139 points are labeled as straight ahead, 7459 points are from right turns and 23953 points are left turns points.
For Sensor 3 we analyse 80 observed trajectories and select 20 of them randomly.
For Sensor 4, 50 out of 146 trajectories are randomly chosen and used for the registration.
The qualitative result is shown in Fig. \ref{fig:Result}.
We see that all trajectories are aligned into the lanes, indicating an acceptable accuracy.
We further show the average mean error in Table \ref{tab:Result}.

All results are in sub-meter range. 
The mean 2D errors for the sensors are 0.66m,  0.94m, 0.38m, 0.55m, respectively.
We also see the results are better in the real world than in the simulation.
The reason of this could be that the real world road has more curvatures for the SICP algorithm to consider, whereas the road in the CARLA map are simpler in its topology.
The minimum error for one axis could reach values below 20 cm. 
The error in yaw angle estimation is also less than 0.4° for both simulation and real-world scenarios.
Overall, with this accuracy, ITS roadside radar sensors could work fine for down stream tasks like object tracking and traffic analysis.
\subsection{Comparison with other method}
We compare the result of localizing S3 and S4 with the method proposed in \cite{localizationiv}.
The difference of these two scenes lies mainly in the size of the observing areas.
Sensor 3 monitors close to a intersection, whereas Sensor 4 observes vehicles in a larger field.
The translational error in both $x$ and $y$ direction are reduced using this new method. 
For Sensor 3, the yaw angle error is also reduced using the new method.
This shows that by analyzing the trajectories, the point cloud registration is done in a more comprehensive manner. 
However, we observe a slight increase in the yaw error for Sensor 4 when applying the new methods.
We argue the reason lies in possible different data patterns from the accumulated dynamic point clouds in \cite{localizationiv} and the observed trajectories.
This could result from the number of the seen vehicles.
Comparing to the large number of points in 8000 frames used in \cite{localizationiv}, the new method achieves great data efficiency.
Given the fact that object tracking is a main task for ITS roadside sensors and needs to be conducted anyway, this methods further saves computational resources.

\begin{table}
    \centering
    \caption{Comparison between two ITS sensor localization methods}
    \label{tab:Comparison}
    \begin{tabular}{cccc}
        \hline
         Sensor&  $\text{Error}_x$ (m)&  $\text{Error}_y$ (m)& $\text{Error}_{yaw}$ (°)\\
        \hline
        S3 \cite{localizationiv}&  0.93&  0.82& 1.25\\
         S3 (ours)&  \textbf{0.32}&  \textbf{0.21}& \textbf{0.35}\\
         S4 \cite{localizationiv}& 0.80& 0.70&\textbf{0.03}\\
         S4 (ours)& \textbf{0.52}& \textbf{0.19}&0.16\\
    \end{tabular}
\end{table}

\section{Conclusion}
\label{Chapter6: Conclusion}

We have proposed a novel self-localization method for ITS roadside radar sensor employing Extended Object Tracking.
The method analyses both the tracked trajectories of the vehicles observed by the sensor and the aerial laser scan of the city streets, assigns labels of driving behaviors like "straight ahead", "left turn", "right turn", and performs SICP algorithm to register the point cloud.
The result of SICP is the transformation matrix for the localization task in a global reference frame. 
The required combination of algorithms is illustrated and comprehensively described, which is able to be automated.
Evaluations are shown in both simulation and real-world scenarios.
The method is demonstrated to deliver accurate sensor localization results with high data efficiency.
Further development for the localization task lies in the extending the current 2D EOT method to a 3D version, and improving the state estimation by replacing the filter with a smoother. 
Another possible exploitation is the generation of a data-based lane description using the tracked trajectories, for example at a junction to study the true driving behaviors when lane markings are not explicit.

\bibliographystyle{IEEEtran} 
\bibliography{refs} 

\end{document}